\documentclass[11pt]{article}
\usepackage[utf8]{inputenc}
\usepackage{nodalida2021}
\usepackage{times}
\usepackage{url}
\usepackage{latexsym}
\usepackage{appendix}
\usepackage{xspace}
\usepackage{todonotes}
\usepackage{hyperref}
\usepackage{verbatim}
\usepackage{multirow}
\usepackage{booktabs}
\usepackage{covington}
\usepackage{comment}
\usepackage{graphicx}

\newcommand{\ie}{\textit{i.e.}\xspace}

\newcommand{\eg}{\textit{e.g.}\xspace}
\newcommand{\F}{F$_1$\xspace}

\newcommand{\emoji}{
    \includegraphics[scale=0.2]{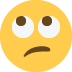}
}

\definecolor{lightred}{RGB}{255,220,220}
\newcommand{\trait}[1]{{\setlength{\fboxsep}{3pt}\colorbox{lightred}{#1}}}

\aclfinalcopy 

\title{NorDial: A Preliminary Corpus of Written Norwegian Dialect Use}

 \author{Jeremy Barnes\thanks{~~The authors have equal contribution.} \\
   Department of Informatics \\
   University of Oslo \\
   {\tt jeremycb@uio.no} \\ \And
   Petter Mæhlum\footnotemark[1] \\
   Department of Informatics \\
   University of Oslo \\
   {\tt petterma@uio.no} \\ \And
   Samia Touileb\footnotemark[1] \\
   Department of Informatics \\
   University of Oslo \\
   {\tt samiat@uio.no} \\
   }

\date{}

\begin{document}
\maketitle
\begin{abstract}
Norway has a large amount of dialectal variation,
as well as a general tolerance to its use in the public sphere. There are, however, few available resources to study this variation and its change over time and in more informal areas, \eg on social media. In this paper, we propose a first step to creating a corpus of dialectal variation of written Norwegian. We collect a small corpus of tweets and manually annotate them as Bokmål, Nynorsk, any dialect, or a mix. We further perform preliminary experiments with state-of-the-art models, as well as an analysis of the data to expand this corpus in the future. Finally, we make the annotations and models available for future work.

\end{abstract}

\section{Introduction}


Norway has a large tolerance towards dialectal variation \cite{nhs} and, as such, one can find examples of dialectal use in many areas of the public sphere, including politics, news media, and social media. %
Although there has been much variation in writing Norwegian, since the debut of Nynorsk in the 1850's, the acceptance of dialect use in certain settings is relatively new. The official language policy after World War 2 was to include forms belonging to all layers of society into the written norms, and a ``dialect wave'' has been going on since the 1970's  \cite[235-238]{nhs}. 

From 1980 to 1983 there was an ongoing project called \textit{Den første lese- og skriveopplæring på dialekt} `The first training in reading and writing in dialect' \cite{lesing-og-barns-talemaal-bull}, where primary school students were allowed to use their own dialect in school, with Tove Bull as project leader. \newcite{nhs} also point out that later interest in writing in dialect in media such as e-mail and text messages can be seen as an extension of the interest in dialectal writing in the 1980s \cite[239]{nhs}. They also note that the tendency has been the strongest in the county of Trøndelag initially, but later spreading to other parts of the country, also spreading among adults. 

At the same time, there are two official main writing systems, \ie Bokmål and Nynorsk, which offer prescriptive rules for how to write the spoken variants. This leads to a situation where people who typically use their dialect when speaking often revert to one of the written standards when writing. However, despite there being only two official writing systems, there is considerable variation within each system, as the result of years of language policies. Today we can find both `radical' and `conservative' versions of each writing system, where the radical ones try to bridge the gap between the two norms, while the conservative versions attempt to preserve differences. However, it is still natural that these standards have a regularizing effect on the written varieties of people who normally speak their dialect in most situations \cite{Gal2017}. As such, it would be interesting to know \emph{to what degree dialect users deviate from these established norms and use dialect traits when writing informal texts}, \eg on social media. This could also provide evidence of the vitality of certain dialectal traits.

In this paper, we propose a first step towards creating a corpus of written dialectal Norwegian by identifying the best methods to collect, clean, and annotate tweets into Bokmål, Nynorsk, or dialectal Norwegian. We concentrate on geolects, rather than sociolects, as we observe these are easier to collect on Twitter, \ie the traits that identify a geolect are more likely to be written than those that identify a sociolect. This is a necessary simplification, as dialect users rarely write with full phonetic awareness, making it impossible to find dialect traits that lie mainly in the phonology. As such, our corpus is relies more on lexical and clear phonetic traits to determine whether a tweet is written in a dialect.

We collect a corpus of 1,073 tweets which are annotated as \texttt{Bokmål}, \texttt{Nynorsk}, \texttt{Dialect}, or \texttt{Mixed} and perform a first set of experiments to classify tweets as containing dialectal traits using state-of-the-art methods. We find that fine-tuning a Norwegian BERT model (NB-BERT) leads to the best results. We perform an analysis of the data to find useful features for searching for tweets in the future, confirming several linguistic observations of common dialectal traits and find that certain dialectal traits (those from Trøndelag) are more likely to be written, suggesting that since their traits strongly diverge from Bokmål and Nynorsk, they are more likely to deviate from the established norms when composing tweets. Finally, we release the annotations and dialect prediction models for future research.\footnote{Available at \url{https://github.com/jerbarnes/norwegian_dialect}}










\section{Related Work}

The importance of incorporating language variation into natural language processing approaches has gained visibility in recent years. The VarDial workshop series deals with computational methods and language resources for closely related languages, language varieties, and dialects and have offered shared tasks on language variety identification for Romanian, German, Uralic languages \cite{zampieri-etal-2019-report}, among others. Similarly, there have been shared tasks on Arabic dialect identification \cite{bouamor2019madar, abdulmageedetal2020nadi}. To our knowledge, however, there are no available written dialect identification corpora for Norwegian.

Many successful approaches to dialect identification use linear models (\eg Support Vector Machines, Multinomial Naive Bayes) with word and character n-gram features \cite{wu-etal-2019-language,jauhiainen-etal-2019-discriminating}, while neural approaches often perform poorly \cite{zampieri-etal-2019-report} (see \citet{jauhiainen-etal-2019} for a full discussion). More recent uses of pretrained language models based on transformer architectures \cite{devlin-etal-2019-bert}, however, have shown promise \cite{bernier-colborne-etal-2019-improving}.

Corpus-related work on Norwegian dialects has mainly focused on spoken varieties. There are two larger corpora available for Norwegian: the newer Nordic Dialect Corpus \cite{johannessen-etal-2009-nordic}, which contains spoken data from several Nordic languages, and the Language Infrastructure made Accessible (LIA) Corpus, which in addition to Norwegian also contain S\'{a}mi language clips.\footnote{\href{https://www.hf.uio.no/iln/english/research/projects/language-infrastructure-made-accessible/}{https://www.hf.uio.no/iln/english/research/projects/language-infrastructure-made-accessible/}} There is also the \textit{Talk of Norway} Corpus \cite{lapponi-etal-2018}, which contains transcriptions of parliamentary speeches in a variety of language varieties. While they contain rich dialectal information, this information is not kept in writing, as they are normalized to Bokmål and Nynorsk. These resources are useful for working with speech technology and questions about Norwegian dialects as they are spoken, but they are likely not sufficient to answer research questions about how dialects are expressed when written. The transcriptions in these corpora also differ from written dialect sources in the sense that they are in a way truer representations of the dialects in question. In writing dialect representations tend to focus more on a few core words, even if the actual phonetic realization of certain words could have been marked in writing.

\section{Data collection}


In this first round of annotations, we search for tweets containing Bokmål, Nynorsk, and Dialect terms (See Appendix \ref{appendix}), discarding tweets that are shorter than 10 tokens. The terms were collected by gathering frequency bigram lists from the Nordic Dialect Corpus \cite{johannessen-etal-2009-nordic} from the written representation of the dialectal varieties.

Two native speakers annotated these tweets with four labels: \texttt{Bokmål}, \texttt{Nynorsk}, \texttt{Dialect}, and \texttt{Mixed}. The \texttt{Mixed} class refers to tweets where there is a clear separation of dialectal and non-dialectal texts, \eg reported speech in \texttt{Bokmål} with comments in \texttt{Dialect}. This class can be very problematic for our classification task, as the content can be a mix of all the other three classes. We nevertheless keep it, as it still reflects one of the written representations of Norwegian.  

In Example \ref{ex:nordiaexample}, we show two phrases from the Nordic Dialect Corpus, from a speaker in Ballangen, Nordland county. We show it in dialectal form (a) and the Bokmål (b) transcription, but with added punctuation marks. To exemplify the two other categories we have manually translated it to Nynorsk (c) and added a mixed version (d), as well as an English translation (e) for reader comprehension.


\begin{table}[]
    \centering
    \resizebox{.45\textwidth}{!}{
    \begin{tabular}{lrrrrrrrrrrrr}
    \toprule
    & Bokmål & Nynorsk & Dialect & Mixed & \textbf{Total
}\\ 
    \cmidrule(lr){2-2}\cmidrule(lr){3-3}\cmidrule(lr){4-4}\cmidrule(lr){5-5}\cmidrule(lr){6-6}
    Train &  348 & 174 & 274 & 52 & 848 \\
    Dev &  52 & 20 & 30 & 4 & 106\\
    Test &  38 & 31 & 35 & 6 & 110\\
    \cmidrule(lr){2-2}\cmidrule(lr){3-3}\cmidrule(lr){4-4}\cmidrule(lr){5-5}\cmidrule(lr){6-6}
    \textbf{Total} & 438 & 225 & 348 & 62 & 1,073 \\
    \bottomrule
    \end{tabular}
     }
    \caption{Data statistics for the corpus, including number of tweets per split.}
    \label{tab:stats}
\end{table}

\begin{covexample}
\begin{itemize}
    \item[(a)] Æ ha løsst å fær dit. Æ har løsst å gå på skole dær.\\
    \item[(b)] Jeg har lyst å fare dit. Jeg har lyst å gå på skole der.\\
    \item[(c)] Eg har lyst å fara dit. Eg har lyst å gå på skule der.\\
    \item[(d)] Æ ha løsst å fær dit. Jeg har lyst å gå på skole der. \\
    \item[(e)] I want to go there. I want to go to school there.
\end{itemize}
\label{ex:nordiaexample}
\end{covexample}

The two annotators doubly annotated a subset of the data in order to assess inter annotator agreement. On a subset of 126 tweets, they achieved a Cohen's Kappa score of 0.76, which corresponds to substantial agreement. Given the strong agreement on this subset, we did not require double annotations for the remaining tweets. Table \ref{tab:stats} shows the final distribution of tweets in the training, development, and test splits. \texttt{Bokmål} tweets are the most common, followed by \texttt{Dialect} and \texttt{Nynorsk}, and as can be seen, \texttt{Mixed} represents a smaller subset of the data.

Certain traits made the annotation difficult. Many tweets, especially those written in dialect, are informal, and therefore contain more slang and spelling mistakes. For example, \textit{jeg} `I' can be misspelled as \textit{eg}, which if found in a non-Nynorsk setting could indicate dialectal variation. Spelling mistakes should not interfere with dialect identification, but as some tweets can contain as little as one token that serve to identify the language variety as dialectal, this can cause problems. Some dialects are also quite similar to either Bokmål or Nynorsk, and speakers might switch between them when speaking or writing.  
Similarly, certain elements can be indicative of either a geolect or a sociolect, \eg the pronoun \textit{dem} `they' as the third person plural subject pronoun (\textit{de} in Bokmål and Nynorsk), which in a rural setting might be typical for an East Norwegian dialect, while in an urban setting might be a strong sociolectal indicator. Tweets with similar problems are annotated in favor of the dialect class. Additionally, there is the problem of internal variation. A tweet can belong to a radical or conservative variety of standardized Norwegian, \eg Riksmål, and thereby not be dialectal. However, this distinction can be difficult to make if a writer uses forms that are now removed from the main standards (Bokmål and Nynorsk), and therefore become more marked, such as \textit{sprog} instead of \textit{språk} `language'.

\section{Dialectal traits}

\begin{table}[]
    \centering
    \begin{tabular}{lrlr}
    \toprule
 \multicolumn{2}{c}{Bokmål-Dialect} & \multicolumn{2}{c}{Nynorsk-Dialect} \\
 \cmidrule(lr){1-2}\cmidrule(lr){3-4}
 `e' & 288.7 & `e' & 131.8\\
 `æ' & 188.0 & `æ' & 92.5\\
 `ska' & 55.0 & `ska' & 23.9\\
 `hu' & 36.6 & `ei' & 18.9\\
 `te' & 28.9 & `berre' & 14.5\\
 (`æ', `e') & 27.5 & `hu' & 14.4\\
 `ka' & 22.0 & `heilt' & 13.8\\
 `mæ' & 21.6 & (`æ', `e') & 13.2\\
 `går' & 19.9 & `meir' & 12.3\\
 `va' & 12.4 & `mæ' & 11.9\\
\bottomrule
    \end{tabular}
    \caption{Top 10 features and $\chi^{2}$ values between Bokmål -- Dialect tweets and Nynorsk -- Dialect.}
    \label{tab:chisquared}
\end{table}

To find the most salient written dialect traits compared to Bokmål and Nynorsk, we perform a $\chi^{2}$ test \cite{Pearson1900} on the occurrence of unigrams, bigrams, and trigrams pairwise between Bokmål and Dialect, and then Nynorsk and Dialect and set $p = 0.5$. 

The most salient features (see Table \ref{tab:chisquared}) are mainly unigrams that contain dialect features, \eg \textit{æ} `I', \textit{e} `am/is/are', \textit{ska} `shall/will', \textit{te} `to', \textit{mæ} `me', \textit{frå} `from', although there are also two statistically significant bigrams, \eg \textit{æ e} `I am', \textit{æ ska} `I will'.
We notice that many of these features likely correspond to Trøndersk and Nordnorsk variants. Similar features from other dialects (\textit{i}, \textit{jæ}, \textit{je} `I') are not currently found in the corpus. This may reflect the natural usage, but it is also possible that the original search query should be improved. Example \ref{ex:dialect_example} shows an example of a \texttt{Dialect} tweet (the English translation is 'Now you know how I've felt for a few years') where the dialectal words have been highlighted.

\begin{covexample}
\texttt{Nå vet du \trait{åssen æ} har hatt det i noen år \emoji} 
\label{ex:dialect_example}
\end{covexample}

\section{Experiments}

We propose baseline experiments on a 80/10/10 split for training, development and testing and use a Multinomial Naive Bayes (MNB) and a linear SVM. As features, we use tf–idf word and character (1-5) n-gram features, with a minimum document frequency of 5 for words, and 2 for characters. We use MNB with alpha=0.01, and SVM with hinge loss and regularization of 0.5 and use grid search to identify the best combination of parameters and features.   

We also compare two Norwegian BERT models: NorBERT\footnote{\url{https://huggingface.co/ltgoslo/norbert}} \cite{KutBarVel21} and NB-BERT\footnote{\url{https://huggingface.co/NbAiLab/nb-bert-base}} \cite{Kummervold2021}, 
which use the same architecture as BERT base cased \cite{devlin-etal-2019-bert}. NorBERT uses a 28,600 entry Norwegian-specific sentence piece vocabulary and was jointly trained on 200M sentences in Bokmål and Nynorsk, while NB-BERT uses the vocabulary from multilingual BERT and is trained on 18 billion tokens from a variety of sources\footnote{See \url{https://github.com/NBAiLab/notram}.}, including historical texts, which presumably contain more examples of written dialect. 
We use the huggingface transformers implementation and feed the final `[CLS]' embedding to a linear layer, followed by a softmax for classification. The only hyperparameter we optimize is the number of training epochs. We use weight decay on all parameters except for the bias and layer norms and set the learning rate for AdamW \cite{loshchilov2018decoupled} to $1e-5$ and set all other hyperparameters to default settings. We train the model for 20 epochs, and keep the model that achieves the best macro \F on the dev set. 

\begin{table}[t]
    \centering
    \begin{tabular}{llrrr}
    \toprule
    & & Precision & Recall & \F \\
    \cmidrule(lr){2-2}\cmidrule(lr){3-3}\cmidrule(lr){4-4}\cmidrule(lr){5-5}
    \multirow{4}{*}{\rotatebox{90}{DEV}} &
     MNB    & 0.70 & 0.67 & 0.68 \\
     &
     SVM     & 0.87 & 0.69 & 0.73\\
     &
     NorBERT & 0.73 & 0.72 & 0.72 \\
     &
     NB-BERT & \textbf{0.89} & \textbf{0.90} & \textbf{0.89} \\
     \cmidrule(lr){2-2}\cmidrule(lr){3-3}\cmidrule(lr){4-4}\cmidrule(lr){5-5}
     \multirow{4}{*}{\rotatebox{90}{TEST}} &
     MNB  & 0.60 & 0.61 & 0.60  \\
     &
     SVM   & \textbf{0.86} & 0.67 & 0.69  \\
     &
     NorBERT & 0.73 & 0.72 & 0.72\\
     & 
     NB-BERT & 0.81 & \textbf{0.78} & \textbf{0.79} \\
     \bottomrule
    \end{tabular}
    \caption{Precision, recall, and macro \F for each model, on the dev and test sets.}
    \label{tab:results}
\end{table}

Table \ref{tab:results} shows the results for all models. MNB is the weakest model on both dev and test on all metrics. Despite the fact that it usually gives good results for dialect identification, it is quite clear that it does not fit our dataset. We think that this might mainly be due to the large vocabulary overlap between the datasets, especially in the \texttt{Mixed} class. SVM has the best precision on both dev (0.87) and test (0.86) and the best \F on dev, while recall on each is lower (0.69/0.67). NB-BERT has the best recall on both dev and test, and is the best overall model on \F (0.79), followed by NorBERT.

\section{Error analysis}

\begin{figure}[t]
    \centering
    \includegraphics[width=.45\textwidth]{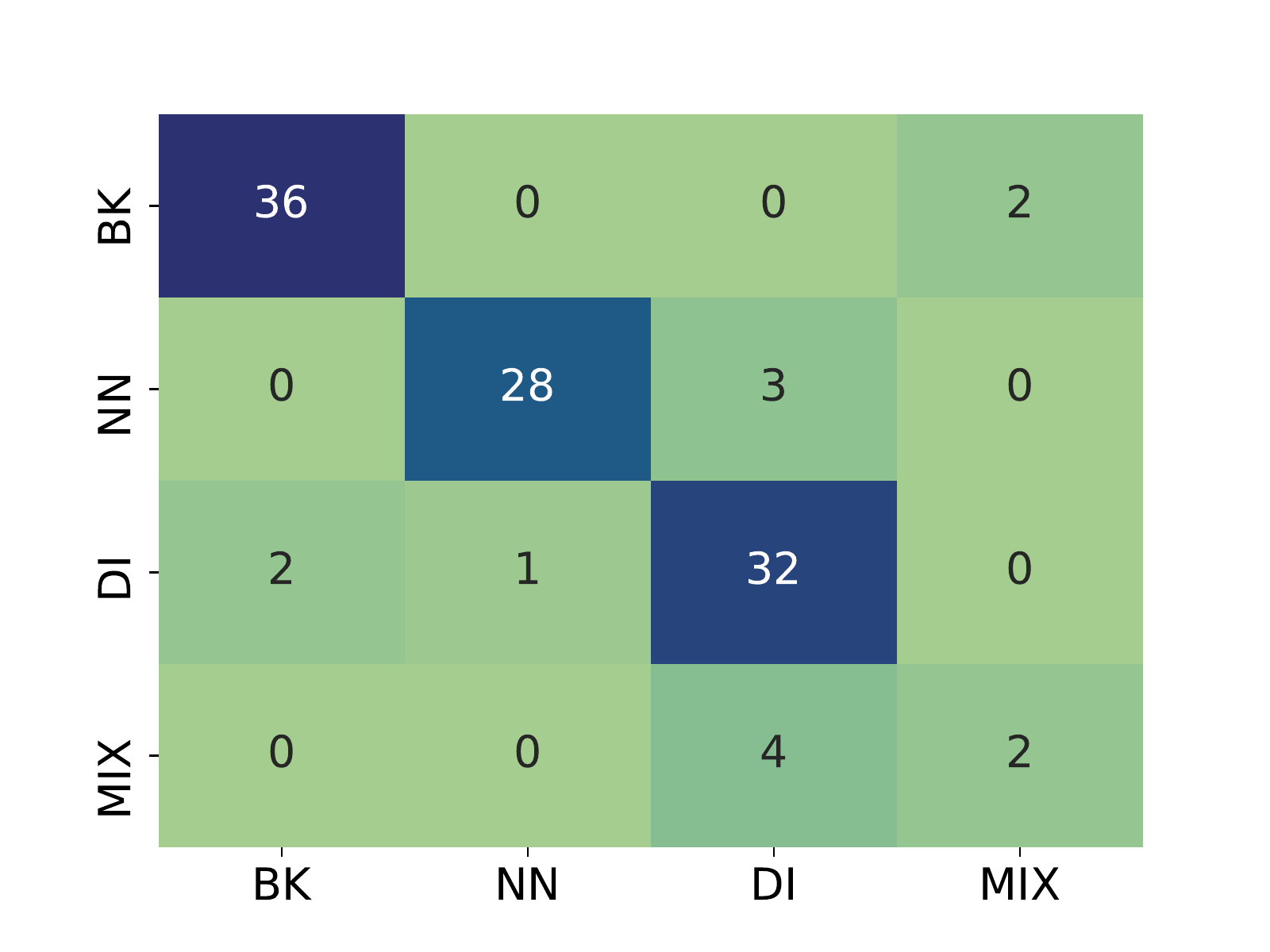}
    \caption{Confusion matrix of NB-BERT on Bokmål (BK), Nynorsk (NN), Dialect (DI), and Mixed (MIX). }
    \label{fig:confusion}
\end{figure}

Figure \ref{fig:confusion} 
shows 
a confusion matrix of NB-BERT's predictions on the test data. The main three categories (\texttt{Bokmål}, \texttt{Nynorsk}, and \texttt{Dialect}) are generally well predicted, while \texttt{Mixed} is currently the hardest category to predict. This is expected, as the \texttt{Mixed} class comprises all of the three other forms. The model has a tendency to predict \texttt{Nynorsk} or \texttt{Mixed} for \texttt{Dialect} and struggles with \texttt{Mixed}, predicting either \texttt{Bokmål} or \texttt{Dialect}. The same observations apply to NorBERT, MNB, and SVM classifiers. 

Given that our main interest lies in the ability to predict future \texttt{Dialect} tweets, we compute precision, recall, and \F on only this label. The NB-BERT model achieves 0.82, 0.91, and 0.86, respectively while NorBERT follows with 0.84, 0.77, and 0.81. The SVM model achieves 0.80, 0.69, and 0.74 respectively, while MNB obtains slightly less scores with respectively 0.77, 0.66, and 0.71. This suggests that future experiments should consider using NB-BERT.

\section{Conclusion and Future Work}

In this paper we have described our first annotation effort to create a corpus of dialectal variation in written Norwegian. In the future, we plan to use our trained models to expand the corpus in a semi-supervised fashion by refining our searches for tweets with dialectal traits in order to have a larger corpus of dialectal tweets, effectively pursuing a high-precision low-recall path. In parallel, we will begin to download large numbers of tweets and use our trained models to automatically annotate these (low-precision, high-recall). 
At the same time we plan to perform continuous manual evaluations of small amounts of the data in order to identify a larger variety of dialectal tweets, which we will incorporate into the training data for future models. 

Second, we would like to annotate these dialectal tweets with their specific dialect. To avoid collecting too many tweets from overrepresented dialects, we will first annotate the current dialectal tweets with their dialect, and perform a balanced search to find a similar number of tweets for each dialect.

Finally, we would like to incorporate texts from different sources which contain rich dialectal variation, as \eg books, music, poetry.


\bibliographystyle{acl_natbib}
\bibliography{nodalida2021}

\clearpage
\appendix
\section{Appendix}
\label{appendix}

\textbf{Bokmål terms:} `jeg har', `de går', `jeg skal', `jeg blir', `de skal', `jeg er', `de blir', `de har', `de er', `dere går', `dere skal', `dere blir', `dere har', `dere er', `hun går', `hun skal', `hun blir', `hun har', `hun er', `jeg går'.

\textbf{Nynorsk terms:} `eg har', `dei går', `eg skal', `eg blir', `dei skal', `eg er', `dei blir', `dei har', `dei er', `de går', `dykk går','de skal','dykk skal','de blir','dykk blir','de har','dykk har','de er','dykk er', `ho gaar', `ho skal', `ho blir', `ho har', `ho er', `eg går'.

\textbf{Dialect terms:} `e ha', `æ ha', `æ har', `e har', `jæ ha', `eg har', `eg ha', `je ha', `jæ har', `di går', `demm går', `dem går', `dæmm går', `dæm går', `dæi går', `demm gå', `dem gå', `di går', `domm gå', `dom gå', `dømm går', `døm går', `dæmm gå', `dæm gå', `e ska', `æ ska', `jæ ska', `eg ska', `je ska', `i ska', `ei ska', `jæi ska', `je skæ', `e bli', `æ bli', `jæ bli', `e bi', `æ blir', `æ bi', `je bli', `e blir', `i bli', `di ska', `dæmm ska', `dæm ska', `dæi ska', `demm ska', `dem ska', `domm ska', `dom ska', `dømm ska', `døm ska', `dæ ska', `domm ska', `dom ska', `æmm ska', `æm ska', `eg e', `æ e', `e e', `jæ æ', `e æ', `jæ ær', `je æ', `i e', `æg e', `di bi', `di bli', `dæi bli', `dæmm bli', `dæm bli', `di blir', `demm bli', `dem bli', `dæmm bi', `dæm bi', `dømm bli', `døm bli', `dømm bi', `døm bi', `di har', `di ha', `dæmm ha', `dæm ha', `dæmm har', `dæm har', `dæi he', `demm har', `dem har', `demm ha', `dem ha', `dæi ha', `di he', `dæmm e', `dæm e', `di e', `dæi e', `demm e', `dem e', `di æ', `dømm æ', `døm æ', `demm æ', `dem æ', `dei e', `dæi æ', `dåkk går', `dåkke går', `dåkke gå', `de går', `dåkk ska', `dere ska', `dåkker ska', `dåkke ska', `di ska', `de ska', `åkk ska', `røkk ska', `døkker ska', `døkk bli', `dåkker bi', `dåkke bli', `dåkker har', `dåkker ha', `dere ha', `dåkk ha', `de har', `dåkk har', `dere har', `de ha', `døkk ha', `dåkker e', `dåkk e', `dåkke e', `di e', `dere ær', `dåkk æ', `de e', `økk e', `døkk æ', `ho går', `hu går', `ho jenng', `ho gjenng', `u går', `o går', `ho jænng', `ho gjænng', `ho jenngg', `ho gjenngg', `ho jennge', `ho gjennge', `ho gå', `ho ska', `hu ska', `a ska', `u ska', `o ska', `hu skar', `honn ska', `ho sjka', `hænne ska', `ho bli', `ho bi', `o bli', `ho blir', `hu bli', `hu bler', `hu bi', `ho bir', `a blir', `ho ha', `ho har', `ho he', `hu har', `hu ha', `hu he', `o har', `o ha', `hu e', `ho e', `hu e', `ho æ', `hu æ', `o e', `hu ær', `u e', `ho ær', `ho er', `e går', `æ går', `eg går', `jæ gå', `jæ går', `æ gå', `jæi går', `e gå'.

\end{document}